\documentclass{article}

\usepackage[preprint]{neurips_2026}

\usepackage[utf8]{inputenc} 
\usepackage[T1]{fontenc}    
\usepackage{hyperref}       
\usepackage{url}            
\usepackage{booktabs}       
\usepackage{amsfonts}       
\usepackage{nicefrac}       
\usepackage{microtype}      
\usepackage{xcolor}         
\usepackage{xspace}
\usepackage{multirow}
\usepackage{algorithm}
\usepackage{algorithmic}
\usepackage{pifont}
\usepackage{makecell}
\usepackage{amsmath,amsthm,amssymb}
\usepackage{color}

\newcommand{\beq}{\begin{equation}}
\newcommand{\eeq}{\end{equation}}
\newcommand{\beqs}{\begin{eqnarray}}
\newcommand{\eeqs}{\end{eqnarray}}
\newcommand{\barr}{\begin{array}}
	\newcommand{\earr}{\end{array}}
\newcommand{\bali}{\begin{aligned}}
	\newcommand{\eali}{\end{aligned}}

\newcommand{\Ac}[0]{\ensuremath{\mathcal{A}} }

\newcommand{\Lc}[0]{\ensuremath{\mathcal{L}} }

\newcommand{\Nc}[0]{\ensuremath{\mathcal{N}} }

\newcommand{\Vc}[0]{\ensuremath{\mathcal{V}} }

\newcommand{\Zc}[0]{\ensuremath{\mathcal{Z}} }

\newcommand{\Rbb}[0]{\ensuremath{\mathbb{R}} }

\newcommand{\ie}[0]{\emph{i.e., }}

\newcommand{\eg}[0]{\emph{e.g., }}

\newcommand{\Mmat}[0]{\ensuremath{{\bf M}} }

\newcommand{\Smat}[0]{\ensuremath{{\bf S}} }

\newcommand{\ta}[0]{\ensuremath{\text{a}} }
\newcommand{\tn}[0]{\ensuremath{\text{n}} }
\newcommand{\tq}[0]{\ensuremath{\text{q}} }

\newcommand{\ov}[0]{\ensuremath{\boldsymbol{o}} }

\newcommand{\qv}[0]{\ensuremath{\boldsymbol{q}} }
\newcommand{\rv}[0]{\ensuremath{\boldsymbol{r}} }

\newcommand{\vv}[0]{\ensuremath{\boldsymbol{v}} }

\newcommand{\xv}[0]{\ensuremath{\boldsymbol{x}} }

\newcommand{\zv}[0]{\ensuremath{\boldsymbol{z}} }

\newcommand{\phiv}[0]{\ensuremath{\boldsymbol{\phi}} }

\newcommand{\argmax}{\operatornamewithlimits{argmax}}

\usepackage{scalerel}
\newcommand\mathbox[1]{\mathord{\ThisStyle{%
			\fboxsep3\LMpt\relax\kern1\LMpt\fbox{$\SavedStyle#1$}\kern1\LMpt}}}

\usepackage{caption}
\title{Value-order Decomposition for Generalist Anomaly Detection}

\author{%
	Miaoyun Zhao \thanks{Equal contribution.} 
	\thanks{Corresponding author.
		Key Laboratory of Social Computing and Cognitive Intelligence (Dalian University of Technology), Ministry of Education, Dalian, China. 
	} \\
	Dalian University of Technology\\
	\texttt{myzhao@dlut.edu.cn} \\
	\AND
	Jing Chen \footnotemark[1] \\
	Dalian University of Technology\\
	\texttt{18042639203@163.com} \\
	\And
	Miaoni Zhao \\
	Xi’an Chang’an Vanke City Primary School\\
	\texttt{15829603162@163.com} \\
	\And
	Qiang Zhang \\
	Dalian University of Technology\\
	\texttt{zhangq@dlut.edu.cn} \\
}

\begin{document}

\maketitle

\begin{abstract}
	Industrial anomaly detection suffers from limited data, making cross-domain generalization particularly challenging.
	Generalist Anomaly Detection (GAD) aims to train a unified model on a source domain that can effectively detect anomalies in unseen target domains.
	In the initial semantic feature space, strong entanglement between anomalies and object categories or defect types hinders effective generalization across domains. Recent works address this issue by projecting features into a residual space; however, such methods primarily increase cross-domain overlap for normal features, while anomalous features remain specific to object categories, defect types and data domains, leading to poor alignment and generalization.
	To address this limitation, we propose Value-order Decomposition (VOD), a simple yet effective technique that bridges \textbf{three types of generalization gaps} across object categories, defect types (including real and synthetic defects), and data domains. 
	VOD disentangles and suppresses object-category-, defect-type-, and domain-specific information, promoting alignment within normal and abnormal samples while preserving their separability, thereby enabling robust generalization across the three gaps.
	Leveraging the strong alignment between real and synthetic defects within the same object, we perform anomaly detection using only normal and synthetic-abnormal reference, and effectively generalize to unseen real defect types. Experiments on diverse industrial and medical benchmarks demonstrate that our method, using a simple cut-and-paste anomaly simulation strategy, achieves strong generalization across the three gaps.
\end{abstract}


\begin{figure}[t]
	\vspace{-0.7 cm}
	\setlength{\abovecaptionskip}{2.0pt}
	\centering
	\includegraphics[width=0.99\columnwidth]{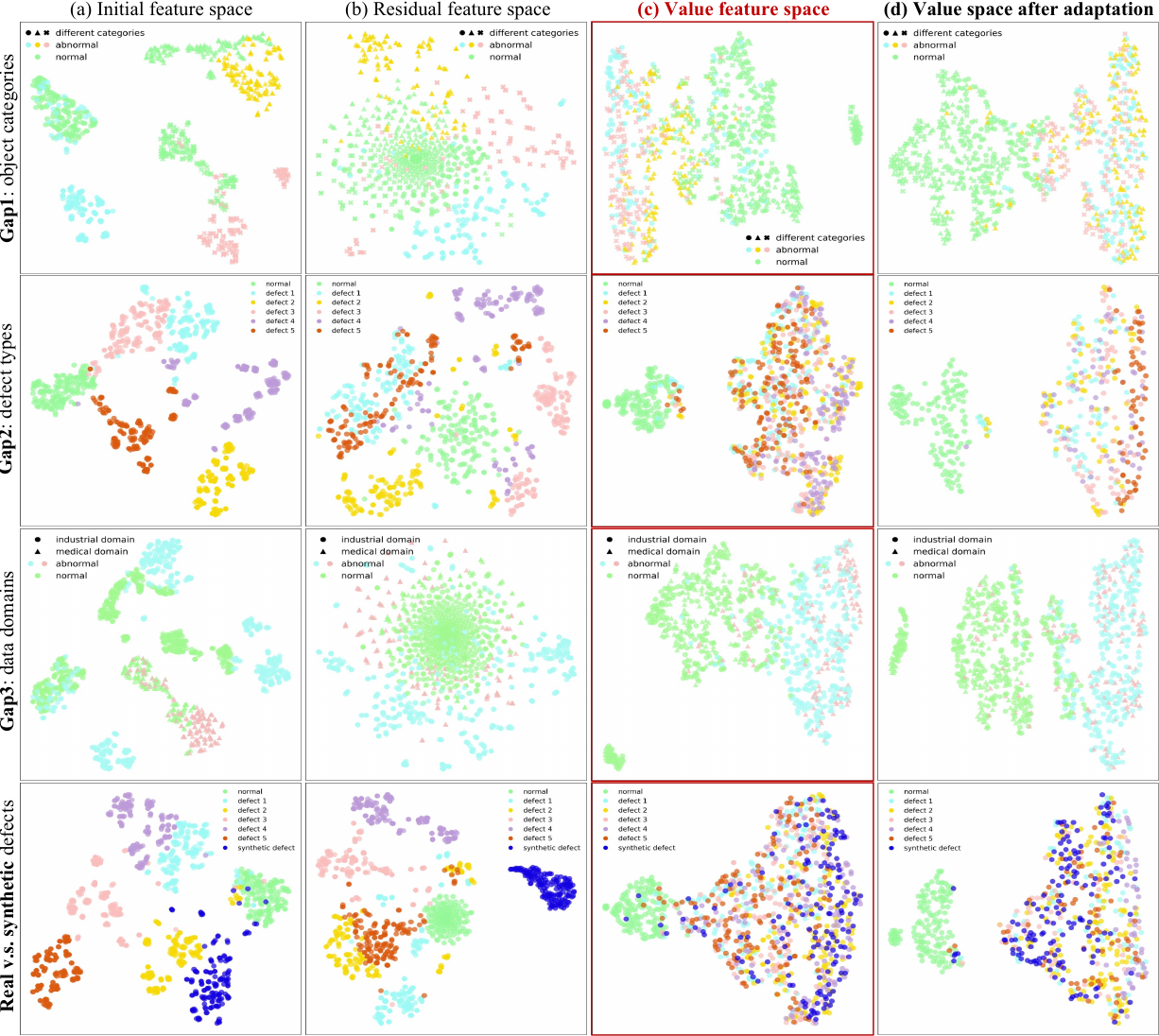}
	\vspace{-0.0 cm}
	\caption{t-SNE visualization of normal and anomalous patches under different representations across the three generalization gaps.
		(a) The initial feature space exhibits three distinct gaps (object categories, defect types, and data domains) rather than the normal/abnormal distinction, which hinders generalization.
		(b) Residual-space methods mainly improve the alignment among normal patches.
		(c) In contrast, the value component of VOD enforces intra-class alignment for both normal and anomalous patches across all gaps.
		(d) After adaptation, the value component further enhances the separability between normal and anomalous patches. (See Appendix \ref{sec:app:full_tsne} for more details.) 
	}\label{fig:three_gaps}
	\vspace{-0.5 cm}
\end{figure}

\section{Introduction}
\label{sec:intro}


Visual anomaly detection (VAD), which aims to classify images as normal or abnormal and localize anomalies, plays a crucial role in industrial quality inspection and medical diagnosis \cite{barusco2025paste,bao2024bmad}.
In real-world applications, however, data scarcity and privacy constraints make it impractical to train a dedicated model for each task. Instead, models are typically pretrained on source domains \cite{yao2024resad, wang2025normalabnormal} and expected to generalize to unseen target domains.
However, VAD is inherently a long-tail problem. Objects and their defects vary significantly in category, color, texture, and size across domains. Consequently, both normal and abnormal features exhibit large variations in the initial feature space (Figure \ref{fig:three_gaps} (a)), leading to insufficiently representative samples and poor generalization of source-trained anomaly detection (AD) models \cite{jeong2023winclip, xu2025towards, gu2025univad}.
To alleviate this issue, a line of unified 
AD methods has been proposed \cite{yao2024hierarchical,yao2023one,yao2023focus,you2022unified}, which retrain or fine-tune models on new classes or domains \cite{yao2024resad}. However, these approaches are computationally expensive and require substantial data, limiting their scalability. As a result, achieving generalization across object categories and domains remains a fundamental challenge in the AD community.

Recently, Generalist Anomaly Detection (GAD) has emerged to address this limitation. It aims to train a unified model on source domains and generalize to unseen target domains without retraining or fine-tuning \cite{yao2024resad,wang2025reinad,zhu2024toward}. Several popular GAD methods focus on learning residual feature distributions rather than initial feature distributions, which effectively reduces variations of normal features across object categories. 
Nevertheless, abnormal residual features still exhibit significant variations across object categories, defect types, and domains. We refer to these variations as the ``\textbf{three generalization gaps}.'' This entanglement hinders generalization to unseen scenarios (see Figure \ref{fig:three_gaps}(b)), which often necessitates elaborate designs \cite{yao2024resad,gao2026adaptclip}. 
To further mitigate this issue, recent works explore normal–abnormal guidance \cite{wang2025normalabnormal}, which leverages anomalous samples of the same object and same defect-type, in addition to normal samples, to better distinguish anomalies across domains. However, such methods rely on predefined defect types, making them infeasible in scenarios where anomaly samples are unavailable, and prone to overfitting to seen defect types, thereby limiting their generalization to novel ones.

To address the three generalization gaps and reduce variations across object categories, defect types, and domains in abnormal features, we focus on the GAD paradigm and propose a novel Value-Order Decomposition (VOD) technique inspired by feature distribution matching \cite{zhang2022exact,rolland2000fast}, enabling strong generalization without requiring abnormal references.
Specifically, VOD decomposes residual features into two components (see Figure \ref{fig:three_gaps} (c) and Figure \ref{fig:tsne_value_order} (a)): ($i$) a value component that captures general anomaly information and remains invariant across object categories, defect types, and domains; 
and ($ii$) an order component that encodes semantic and structural information varying across the three gaps.

Leveraging the value component, we construct an anomaly representation that consistently aligns normal and abnormal samples across diverse object categories, defect types (including real and synthetic ones), and domains. Based on this invariant representation, we further develop a synthetic-anomaly-based detection framework that requires only a few normal samples and generates synthetic anomalies via simple augmentation. During inference, each sample is compared against both normal and synthetic-anomaly references to enable more accurate anomaly detection.
Notably, our method relies only on simple cut-and-paste augmentation, rather than complex diffusion-based synthesis \cite{zhang2024realnet}, while still achieving state-of-the-art generalization performance.
\textbf{The main contributions of this work are:}
($i$) We propose a novel Value-Order Decomposition (VOD) technique that decomposes residual features into gap-invariant and gap-specific components, where the gap-invariant component plays a key role in bridging the three generalization gaps across object categories, defect types, and data domains.
($ii$) By bridging the gap between synthetic and real defects via VOD, we develop a simple yet effective synthetic-anomaly-based framework that avoids complex designs and generalizes well without relying on real anomaly data.
($iii$) Extensive experiments demonstrate that our method, despite its simplicity, outperforms existing baselines by a large margin in most cases, achieving strong generalization and even matching or surpassing methods that rely on real anomaly references.

\section{Related work}
\label{sec:relate}

\subsection{Generalist Anomaly Detection}

Due to the scarcity and high cost of collecting anomalous data, most prior works focus on unsupervised AD, which can be broadly categorized into three technical streams:
($i$) embedding distance-based methods, which identify anomalies through statistical deviations in offline features extracted from pretrained models that preserve discriminative information (\eg SPADE \cite{cohen2020sub}, PaDiM \cite{defard2021padim}, PatchCore \cite{roth2022towards}, HGAD \cite{yao2024hierarchical}, PyramidFlow \cite{lei2023pyramidflow});
($ii$) reconstruction-based methods, which leverage autoencoders \cite{hou2021divide} or GANs \cite{yan2021learning} to detect anomalies through high reconstruction errors \cite{yao2023one,yao2023focus,xiang2023squid};
($iii$) knowledge distillation-based methods, which detect anomalies by measuring discrepancies between student features and those of a fixed pretrained teacher \cite{cao2023anomaly,wang2021glancing,tien2023revisiting,zhao2023omnial,gao2024learning}.
These methods are typically trained using only normal data and identify anomalies by measuring deviations from learned normal patterns. While effective in category-specific settings, they tend to overfit to seen object categories and often perform poorly on unseen categories and domains. Moreover, their reliance on retraining limits practical applicability when data collection is costly or restricted.

To address these limitations, GAD has recently emerged to study cross-domain anomaly detection \cite{zhu2024toward}. In this work, we follow the GAD setting and focus on generalization to unseen domains without retraining or fine-tuning. Existing GAD methods typically achieve generalization by modeling residuals with respect to normal references. 
For example, InCTRL \cite{zhu2024toward} establishes a baseline for cross-dataset anomaly classification by capturing contextual residuals between query images and normal references, but it focuses on classification and lacks precise localization. ResAD \cite{yao2024resad} and AdaptCLIP \cite{gao2026adaptclip} further leverages residual features with explicit normality constraints or comparative learning between query and normal image prompt to enable both detection and localization. 
However, residual-based methods primarily align normal distributions, while abnormal features still exhibit significant variation across domains.
To improve generalization to unseen defect types, some works introduce anomalous references. For instance, NAGL \cite{wang2025normalabnormal} utilizes real anomalous samples to guide detection, but such approaches rely on the availability of anomaly data and are impractical in many real-world scenarios.



To bypass the need for annotated anomalous data, a line of self-supervised methods generates synthetic anomalies with ground-truth annotations and formulates proxy tasks on these data \cite{liu2023simplenet,chen2024unified}. Their performance largely depends on the diversity and realism of the generated anomalies. 
For example, CutPaste \cite{li2021cutpaste} creates anomalies by transplanting image patches, while NSA \cite{perez2023poisson} employs Poisson image editing for more seamless synthesis. DRAEM \cite{zavrtanik2021draem} leverages external texture datasets to generate diverse anomalies. RealNet \cite{zhang2024realnet} and AnoGen \cite{gui2024few} utilizes diffusion models to synthesize more realistic samples. 
However, these methods heavily rely on the quality and diversity of synthetic data, are typically limited to in-domain settings, and require regeneration and retraining when applied to new tasks.

%
%
%
%
%

\subsection{Feature distribution matching}


In image style transfer, feature distributions are closely related to image styles and can be used to align target and source domains while preserving content information \cite{huang2017arbitrary}. This idea has also been widely used in domain generalization.
Several methods study feature distribution alignment from this perspective \cite{shapira2013multiple,coltuc2006exact,hall2006almost,rolland2000fast}. Among them, Sort-Matching \cite{rolland2000fast} performs exact histogram matching, leading to Exact Feature Distribution Matching (EFDM) \cite{zhang2022exact}. EFDM has been successfully applied to both style transfer and domain generalization by aligning feature statistics across domains.

Our VOD is inspired by EFDM but differs in two key aspects.
($i$) We explicitly decompose features into gap-invariant and gap-specific components, enabling the separation of transferable abnormal patterns from variations induced by object categories, defect types, and data domains. In contrast, EFDM focuses on feature mix-up via value-order recombination as a measure-preserving transformation for data augmentation, without explicit feature disentanglement.
($ii$) Instead of jointly utilizing both components as in EFDM, we identify the value component as the most informative for anomaly detection and discard the order component, resulting in a simpler and more robust formulation.


\section{Our method}
\label{sec:our}

\begin{figure}[t]
	\vspace{-0.7 cm}
	\setlength{\abovecaptionskip}{2.0pt}
	\centering
	\includegraphics[width=0.999\columnwidth]{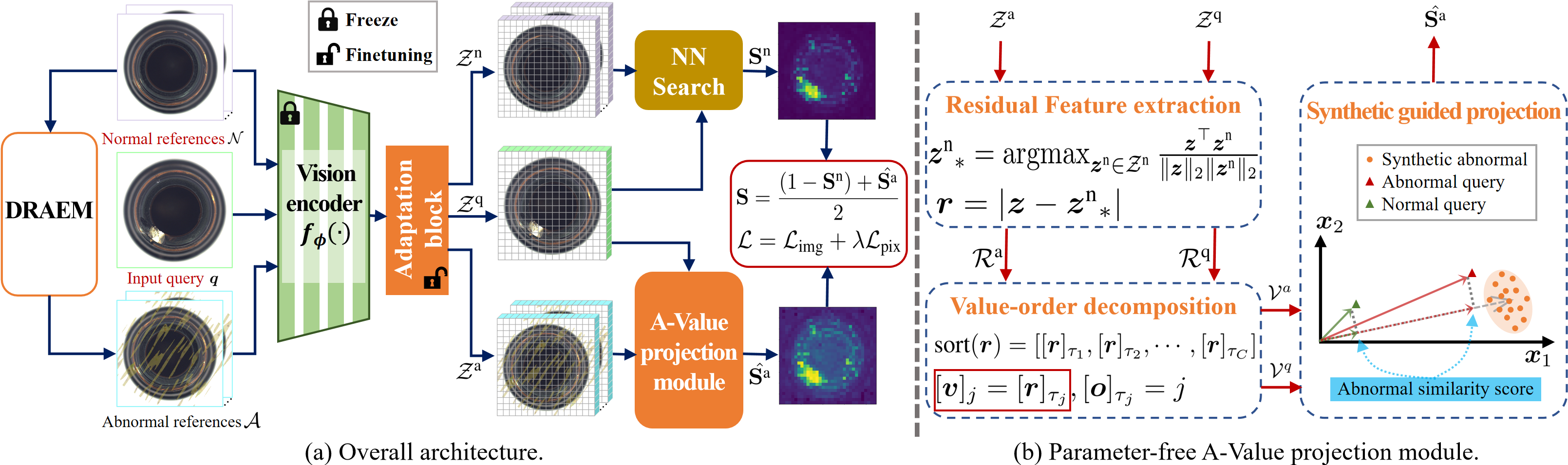}
	\vspace{-0.4 cm}
	\caption{The framework of our VOD-based anomaly detection method. 
	}\label{fig:framework}
	\vspace{-0.5 cm}
\end{figure}

Focusing on the GAD setting, our goal is to train a unified anomaly detection model on a source dataset that generalizes to arbitrary unseen target domains. For a given query image, the detection process requires only a few-shot of normal samples as references.
In the following, we first introduce the proposed VOD technique, highlighting the invariant nature of the value component and the variability captured by the order component.
We then demonstrate the effectiveness of VOD in bridging the three generalization gaps. Based on this, we construct a universal anomaly detection framework that leverages both normal and synthetic-anomaly references.
The overall architecture is illustrated in Figure \ref{fig:framework}.
For a feature vector $\rv\in \mathbb{R}^C$, we use $[\rv]_j$ represents its $j$-th element.

\subsection{Value-order decomposition}

In this section, we introduce the proposed Value-Order Decomposition (VOD) and investigate the strong generalization capability of the value component.
Given a query image $\qv \in \Rbb^{H\times W\times 3}$ and a corresponding normal reference set $\Nc=\{\xv_{k}\}_{k=1}^{K}$ consisting of $K$ shots from the same object category, where $\xv_{k}\in \Rbb^{H\times W\times 3}$, $H$ and $W$ denote the height and width of the image, respectively. 
For both the query and reference images, we extract features using a pre-trained backbone $f_{\phiv}(\cdot)$ parameterized by $\phiv$, following common practice \cite{wang2025normalabnormal,yao2024resad}. 
We then obtain query patch features $\Zc^{\tq}=\{{\zv^{\tq}}_{i}\}_{i=1}^{L}$ and normal reference patch features $\Zc^{\tn}=\{{\zv^{\tn}}_{i}\}_{i=1}^{KL}$, where ${\zv^{\tq}}_{i}, {\zv^{\tn}}_{i}\in \Rbb^{C}$, $L=H\times W$ is the number of patches per image, and $C$ is the channel dimension.

Next, we compute residual features for the query following \cite{yao2024resad} to obtain an initial anomaly localization. Specifically, for each $\zv^{\tq} \in \Zc^{\tq}$, we perform nearest-neighbor search in $\Zc^{\tn}$ using cosine similarity to identify the closest normal reference feature\footnote{Here, we consider Vision Transformer as backbone, where cosine similarity is a more appropriate distance metric.}:
${\zv^{\tn}}_{*}=\argmax_{\zv^{\tn}\in\Zc^{\tn}}\frac{({\zv^{\tq}})^{\top}\zv^{\tn}}{\|{\zv^{\tq}}\|_2\|\zv^{\tn}\|_2}$,
with $\|\cdot\|_2$ denoting the $\ell_2$ norm.
The residual feature is then computed as:
\begin{equation}\label{eq:residual}
	\rv^{\tq} = |\zv^{\tq} - {\zv^{\tn}}_{*} |
\end{equation}
where $|\cdot|$ denotes the element-wise absolute value. By subtracting the most similar normal feature, the residual is expected to suppress object-category-related components in the original features, while preserving the discrepancy between normal and anomalous patterns. 
However, this process primarily aligns the distribution of normal residual features toward an origin-centered region across different object categories, while abnormal residual features remain scattered around the origin with diverse magnitudes and limited separability (Figure \ref{fig:three_gaps} (b)). As a result, a complex one-class classification mechanism is often required to further distinguish normal and abnormal samples \cite{yao2024resad}.

We observe that abnormal residual features tend to cluster according to the three gaps, \ie object categories, defect types, or domains (see Figure \ref{fig:three_gaps} (b)). 
To remove gap-specific information (\eg shapes, colors, and textures) and align the latent representations of anomalous patches, thereby enabling anomaly detection via a simple classification process, we propose Value-Order Decomposition (VOD). 
VOD decomposes residual features into two components: ($i$) a value component that captures gap-invariant information, and ($ii$) an order component that encodes gap-specific structural patterns.
Specifically, we sort the residual feature $\rv^{\tq}$ in descending order:
\begin{equation}\label{eq:sort}
	\text{sort}(\rv^{\tq}) = [[\rv^{\tq}]_{\tau_1}, [\rv^{\tq}]_{\tau_2}, [\rv^{\tq}]_{\tau_3}, \cdots, [\rv^{\tq}]_{\tau_C}] \\
\end{equation}
where $\{[\rv^{\tq}]_{\tau_j}\}_{j=1}^C$ denotes the sorted values of $\rv^{\tq}$, with $[\rv^{\tq}]_{\tau_1}$ being the largest element. Here, $\tau_j$ indicates the index of the $j$-th sorted element in the original feature.
We then define the value component $\vv^{\tq} \in \Rbb^{C}$, where its $j$-th element is computed as:
\begin{equation} \label{eq:value}
	[\vv^{\tq}]_j = [\rv^{\tq}]_{\tau_j}
\end{equation}
Intuitively, the value component $\vv^{\tq}$ corresponds to the sorted version of $\rv^{\tq}$ in descending order, capturing the degree of abnormality (\ie from the most abnormal dimension to the least abnormal one).
The order component $\ov^{\tq} \in \Rbb^{C}$ is then defined such that its element at index $\tau_j$ is given by:
\begin{equation} \label{eq:order}
	[\ov^{\tq}]_{\tau_j} =j
\end{equation}
Specifically, $[\ov^{\tq}]_{\tau_j}$ represents the rank (i.e., the value order) of the element $[\rv^{\tq}]_{\tau_j}$ in the descending order of $\rv^{\tq}$. 
It preserves the relative ordering among different dimensions, thereby retaining defect-type-, object-category-, and domain-specific information. 
For clarity, we provide a toy example: given $\rv^{\tq} = [0.9, 0.6, 0.7, 0.5, 0.8]$, the corresponding VOD results are $\vv^{\tq} = [0.9, 0.8, 0.7, 0.6, 0.5]$ and $\ov^{\tq} = [0, 3, 2, 4, 1]$.

\begin{figure}[t]
	\vspace{-0.5 cm}
	\setlength{\abovecaptionskip}{2.0pt}
	\centering
	\includegraphics[width=0.99\columnwidth]{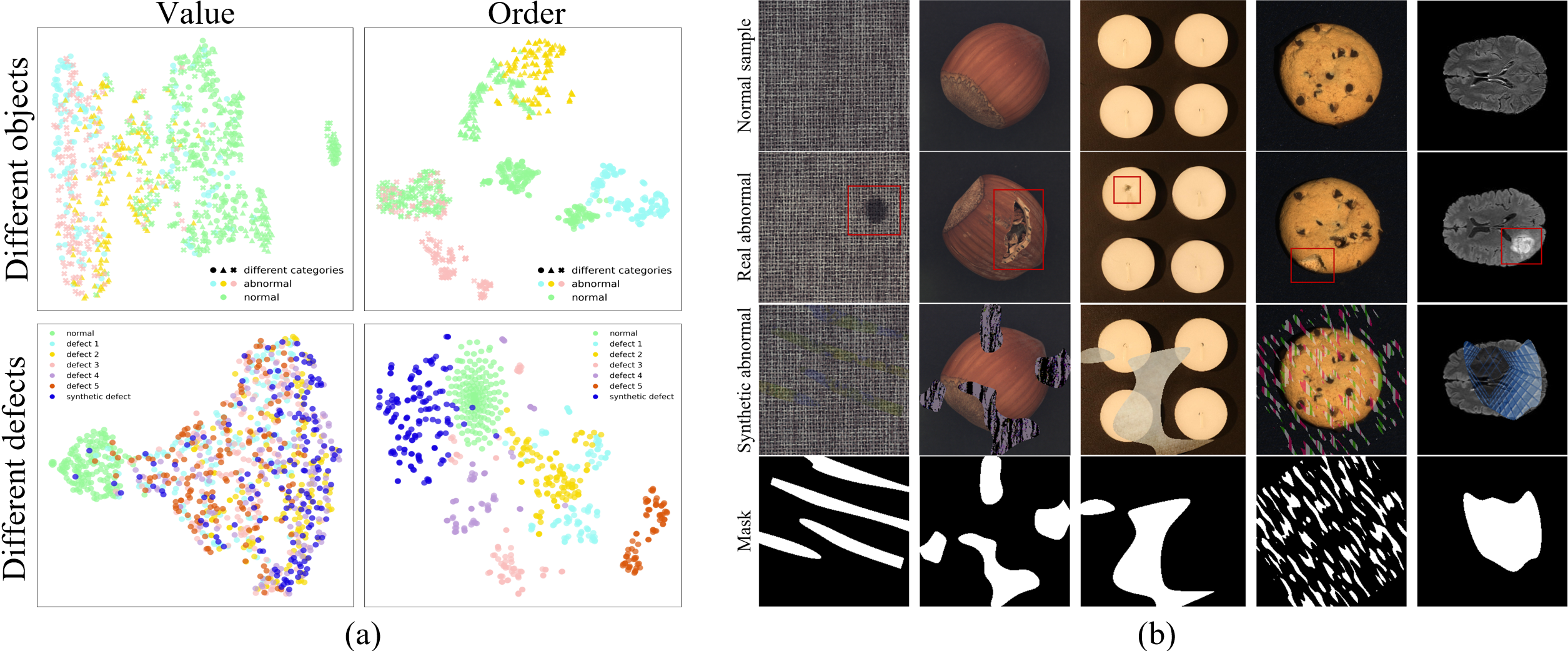}
	\vspace{-0.08 cm}
	\caption{(a) Visualization of the distributions of the value and order components of VOD.
		(b) The synthetic defects used in our work differ from real anomalies.
	}\label{fig:tsne_value_order}
	\vspace{-0.5 cm}
\end{figure}

In this way, the value component captures the element-wise value distribution of $\rv^{\tq}$ without encoding structural information, making it suitable for characterizing abnormality without being affected by the gaps. 
In contrast, the order component records the relative magnitudes of elements (i.e., the structure) without access to their exact values, and is therefore more related to semantics and gap-specific patterns. 
This value-order decomposition implicitly separates gap-invariant and gap-specific information in the residual representation, as empirically demonstrated in Figure \ref{fig:three_gaps}(c) and \ref{fig:tsne_value_order}(a). The order component preserves information related to the three gaps, whereas the value component better aligns distributions across these gaps and exhibits stronger discriminative power between normal and anomalous samples.
Therefore, we adopt $\vv^{\tq}$ as the key representation for anomaly detection.

\textbf{Significance of the Value Component.}
As shown in Figure \ref{fig:three_gaps}(c), beyond bridging the gap among different defect types within the same object category, the value component exhibits stronger invariance properties. In particular, it further reduces the discrepancies across object categories, real and synthetic defect types, and domains.
Specifically:
($i$) the distributions of normal and anomalous features largely overlap across the three gaps;
($ii$) normal and anomalous features become clearly and approximately linearly separable;
($iii$) synthetic defect features effectively capture the mixed distribution of multiple real defect types (last row in Figure \ref{fig:three_gaps}), making them strong proxies for unseen real anomalies and enabling a decision boundary that generalizes well to real anomalies.

The synthetic defects are generated using a simple cut-and-paste anomaly simulation strategy adopted from DRAEM \cite{zavrtanik2021draem}. This approach leverages a Perlin noise generator \cite{perlin1985image} and an external texture dataset \cite{cimpoi2014describing} to produce diverse defects. A few generated defects are shown in Figure \ref{fig:tsne_value_order}(b).
Although the synthetic defects look very different from real ones—the textures are often unrealistic, and the defect regions may spread across both the object and background—our goal is not to faithfully replicate real anomaly appearances in the target domain. Instead, we aim to generate diverse out-of-distribution abnormal patterns that improve generalization.

\subsection{Synthetic-anomaly-based framework for anomaly detection}

Motivated by VOD’s favorable property for synthetic defects, we propose to leverage it for anomaly detection, and \textbf{refer to the overall framework as VOD for simplicity}. The overall pipeline is shown in Figure \ref{fig:framework} and described as follows.

\textbf{Synthetic abnormal reference generation.}
Given $K$-shot normal references, we adopt the synthetic anomaly generation strategy from DRAEM \cite{zavrtanik2021draem}. Specifically, for each normal reference, we generate a random mask $\Mmat$ that captures diverse anomaly shapes. For each mask, only $20\%$ of its values are kept non-zero to define the anomalous regions. We then randomly sample a texture image from DTD \cite{cimpoi2014describing} as the anomaly source. The normal reference is combined with the sampled texture via $\Mmat$, following DRAEM, to produce $K$-shot synthetic abnormal references. 
We then construct a tuple $\{\Nc,\Ac,\qv\}$ consisting of normal references, synthetic abnormal references, and the query image. Anomaly detection is then performed as follows.

We first extract features for each image in the tuple using a pretrained backbone (\eg DINOv2 \cite{oquab2024dinov}) 
with an appended adaptation block (\ie a Transformer encoder layer with 4 attention heads), which provides additional flexibility to enhance the separability between normal and abnormal features. This results in the feature tuple $\{\Zc^{\tn}, \Zc^{\ta}, \Zc^{\tq}\}$.
Based on the assumption that normal patches are mutually similar while abnormal patches exhibit diverse patterns, we compute the anomaly score for each query patch by comparing it with normal references in the original feature space and with abnormal references in the value space of VOD.

\textbf{normal similarity score.}
Given a query patch feature ${\zv^{\tq}}_i\in\Zc^{\tq}$, we perform nearest-neighbor search in $\Zc^{\tn}$ to identify the closest normal feature 
${\zv^{\tn}}_{*}=\argmax_{\zv^{\tn}\in\Zc^{\tn}}\frac{({\zv^{\tq}}_i)^{\top}\zv^{\tn}}{\|{\zv^{\tq}}_i\|_2\|\zv^{\tn}\|_2}$.
Let $\Smat^{\tn}\in\Rbb^L$ denote the normal similarity scores for all query patches. The score for patch ${\zv^{\tq}}_i$ is defined as the cosine similarity to its nearest normal neighbor (rescaled to interval $[0,1]$),
\begin{equation}
	{[\Smat^{\tn}}]_{i} = (\frac{({\zv^{\tq}}_i)^{\top}{\zv^{\tn}}_*}{\|{\zv^{\tq}}_i\|_2\|{\zv^{\tn}}_*\|_2} +1) /2
\end{equation}
\textbf{A-Value projection module for abnormal similarity score.}
To achieve strong generalization, we compute abnormal similarity in the value space. 
We refer to this process as A-Value projection. 
Specifically, we first compute residual features for both abnormal references and the query following Equation \ref{eq:residual}, and then extract the value components via VOD (Equation \ref{eq:value}), resulting in $\Vc^{\ta}=\{{\vv^{\ta}}_{i}\}_{i=1}^{KL}$ and $\Vc^{\tq}=\{{\vv^{\tq}}_{i}\}_{i=1}^{L}$. 
Since real and synthetic abnormal features are well aligned in the value space while remaining clearly separable from normal features, the abnormality of a query feature can be naturally measured by its projection onto abnormal directions (Figure \ref{fig:framework}(b)). The abnormal similarity score $\Smat^{\ta}\in\Rbb^L$ is defined as:  
\begin{equation}
	[{\Smat^{\ta}}]_{i} = \max_{\vv^{\ta}\in\Vc^{\ta}} {\frac{({\vv^{\tq}}_i)^{\top}\vv^{\ta}}{\|\vv^{\ta}\|_2}}
\end{equation}
Here, $\vv^{\ta}\in\Vc^{\ta}$ is normalized to retain only directional information. The anomaly score of ${\vv^{\tq}}_i\in\Vc^{\tq}$ is computed as the maximum projection onto these abnormal directions. The score is further rescaled to $[0,1]$ using the mean norm of synthetic abnormal features:
$[\hat{\Smat^{\ta}}]_{i}=\min([\Smat^{\ta}]_{i}/(\frac{1}{|\Vc^{\ta}|}\sum_{\vv^{\ta}\in\Vc^{\ta}}\|\vv^{\ta}\|_2), 1)$, where values larger than $1$ are clipped to emphasize more challenging anomalies, $|\Vc^{\ta}|$ denotes the number of elements in $\Vc^{\ta}$. 
Finally, the overall anomaly score $\Smat\in\Rbb^L$ is computed by combining the normal similarity and abnormal similarity:
\vspace{-0.2cm}
\begin{equation}
	[\Smat]_{i} = ((1-[\Smat^{\tn}]_{i}) + [\hat{\Smat^{\ta}}]_{i})/2
\end{equation}
\textbf{Training and inference.}
We train VOD in a supervised manner on a source domain and try to generalize to unseen target domains. 
Based on the anomaly score, we adopt the loss function from \cite{wang2025normalabnormal} to jointly optimize image-level and pixel-level predictions,
\begin{equation}\label{eq:loss}
	\Lc = \Lc_\text{img} +\lambda\Lc_\text{pix} \quad
	where \; 
	\Lc_\text{img}= \text{BCE}(s,y), \;
	\Lc_\text{pix} = \text{Focal}(\Smat,\Mmat) + \text{Dice}(\Smat,\Mmat) \\
\end{equation}
Here, $\text{BCE}(\cdot,\cdot)$, $\text{Focal}(\cdot,\cdot)$, $\text{Dice}(\cdot,\cdot)$ denote the binary cross-entropy loss, Focal loss \cite{lin2017focal}, and Dice loss \cite{li2020dice}, respectively. $\lambda$ is used to balance image- and pixel-level losses, which is set to 1 for best performance (see Appendix \ref{sec:app:hyperpara}).
The image-level anomaly score $s$ is defined as the average of the top $1\%$ highest values in $\Smat$. $y$ and $\Mmat$ denote the ground-truth image label and pixel-level anomaly mask, respectively. 

During inference, given a query image and its normal references, we first generate synthetic abnormal references, and then compute both image-level and pix-level anomaly scores as described above to obtain the final detection results. 

%

\section{Experimental results}
\label{sec:Experiments}

We evaluate the effectiveness of the proposed method from three perspectives:
($i$) its ability to generalize across the three generalization gaps: defect types, object categories, and data domains;
($ii$) its plug-and-play compatibility with different backbones, as well as its competitive performance compared to settings that use real abnormal references;
($iii$) an ablation study to analyze the contribution of each technical component.
More experimental results are provided in Appendix \ref{sec:app:moreE}.

\subsection{Experimental Setup}
\label{sec:Experimental_Setup}

\textbf{Datasets.} To demonstrate the broad applicability of our VOD model, we conduct extensive experiments on six datasets spanning both industrial and medical domains.
The industrial domain includes several representative benchmarks, namely MVTecAD \cite{bergmann2019mvtec}, VisA \cite{zou2022spot}, BTAD \cite{9576231}, MVTec3D \cite{bergmann2021mvtec}, and MVTecLOCO \cite{bergmann2022beyond}. In addition, the medical domain is represented by the BraTS \cite{6975210} dataset.


\textbf{Implementation and evaluation metrics.}
We use a DINOv2-S \cite{oquab2024dinov} pre-trained Vision Transformer (ViT) as the feature extractor, with all its parameters frozen while only training the introduced adaptation block. Following prior work \cite{yao2024resad,yao2025resadpp}, we train on one dataset and evaluate on the others. The number of few-shot normal reference samples is set to $K\in\{2,4,8\}$. All experiments are conducted on a single NVIDIA A40 GPU. See Appendix \ref{app:setup} for more implementation details.
Following standard protocols \cite{yao2024resad}, we evaluate anomaly detection at both image and pixel levels using AUROC, where higher values indicate better performance. 





		
%

\textbf{Comparison Methods.} We compare VOD with representative few-shot approaches grouped by the type of reference images used during inference. 
Conventional full-shot AD methods (PaDiM \cite{10.1007/978-3-030-68799-1_35}, PatchCore \cite{roth2022towards}, and SPADE \cite{cohen2020sub}) are adapted to the few-shot setting using a few normal samples for distance-based scoring \cite{yao2024resad}.  
Methods including RegAD\cite{10.1007/978-3-031-20053-3_18},  WinCLIP\cite{Jeong_2023_CVPR}, InCTRL\cite{zhu2024toward}, ResAD\footnote{We adopt the ImageBind-based variant for improved performance.}\cite{yao2024resad}, and AdaptCLIP\cite{gao2026adaptclip} rely only on normal reference images during inference, while NAGL uses both normal and abnormal references.
Among them, AdaptCLIP, ResAD, and NAGL serve as strong baselines, as they demonstrate strong generalization ability across data domains.
 


\subsection{Generalization results across three gaps}
\begin{table}[t]
	\vspace{-0.8 cm}
	\centering
	\caption{
	AUROC (\%) for anomaly detection and localization on 6 real-world datasets under few-shot settings. “·/·” denotes image-/pixel-level results, ``$\Nc$/$\Ac$'' denotes using normal/abnormal references. Values are averaged over 3 runs; best and second-best are bolded and underlined.
		\label{tab:one2many}
	}
	\vspace{-0.03 cm}
	\renewcommand{\arraystretch}{1.15}
	\resizebox{0.99\hsize}{!}{
	\setlength{\tabcolsep}{2pt}
	\begin{tabular}{c|c|cccccccc|c|c}
		\hline
		\multirow{2}{*}{Shots}& \multirow{2}{*}{Datasets} &\multicolumn{8}{c|}{$\Nc$} &
		\multicolumn{1}{c|}{$\Nc$\&$\Ac$} &{$\Nc$}\\
		&  & SPADE & PaDiM & Patchcore & RegAD & WinCLIP & InCTRL&ResAD&AdaptCLIP & NAGL& VOD\\
		\hline
		\multirow{6}{*}{2}& VisA & 71.7/65.4 & 68.7/91.5& 65.0/80.4 & 70.6/93.3 & 81.9/94.9 & 85.8/-& 84.5/95.1 & \underline{92.3}/97.1&89.9/\underline{97.6}& \textbf{92.4}/\textbf{98.4} \\
		& MVTecAD & 74.6/64.0 & 79.5/93.8 & 74.7/85.2 & 80.4/93.3 & 93.1/93.8 & 94.0/- & 94.4/95.6 & \underline{95.7}/94.5 & \textbf{96.8}/\underline{96.8} &\underline{95.7}/\textbf{97.4}  \\
		& BTAD & 80.7/65.4& 88.9/95.2& 80.9/83.1 &87.2/93.9 & 85.5/95.8 & 92.3/- & 91.1/96.4& \underline{93.4}/\textbf{96.7}& 93.0/76.1 & \textbf{95.4}/\underline{96.0}\\
		& MVTec3D & 62.5/78.6 & 59.6/94.3 & 58.8/83.4 &59.5/96.4 & 74.1/96.8 & 68.9/- & 78.5/97.5 & \underline{82.9}/\underline{97.8}& 82.8/94.8 &\textbf{84.2}/\textbf{98.0} \\
		& MVTecLOCO & 59.3/65.7 & 53.2/66.4 &57.5/68.0 &55.0/66.2& 62.6/64.7 & 58.7/-& 65.6/68.0 &\textbf{67.4}/\underline{69.1} & 64.6/66.3& \underline{66.7}/\textbf{75.5}\\
		\cline{2-12}
		&BraTS & 58.0/92.8 & 49.4/90.2 & 58.2/93.5 &54.6/81.4 & 55.6/91.5 & 74.6/-& 67.9/94.3  &\underline{85.7}/\underline{97.1}&82.1/96.8 & \textbf{88.5}/\textbf{97.6}\\
		\hline
		\multirow{6}{*}{4}& VisA & 75.0/65.4 & 75.3/93.3& 71.7/87.1 & 78.0/93.5 & 84.1/95.2 & 87.7/-& 90.8/97.5 &\underline{93.1}/97.3 &91.2/\underline{97.8} & \textbf{93.3}/\textbf{98.5} \\
		& MVTecAD & 75.5/64.0 & 82.5/94.9 & 80.6/90.2 & 84.8/94.5 & 94.6/94.2 & 94.5/-& 94.2/96.9  &96.6/94.8 &\textbf{97.1}/\underline{97.0}  & \underline{96.8}/\textbf{97.5}\\
		& BTAD & 81.7/65.5 & 89.9/\underline{95.8} & 84.0/89.4 &90.8/94.9 & 87.2/\underline{95.8} & 91.7/-& 91.5/\textbf{96.8} &93.3/\textbf{96.8} &93.1/96.1 &\textbf{95.0}/\textbf{96.8}\\
		& MVTec3D & 62.3/78.6 & 62.8/94.5 & 61.5/87.1 &62.3/96.7 & 76.0/97.0& 69.1/- &82.4/97.9 & 84.2/\underline{98.0}& \textbf{86.9}/95.4 & \underline{85.0}/\textbf{98.7} \\
		& MVTecLOCO & 64.0/66.8 & 54.7/67.9 & 61.7/\underline{69.4} &56.6/66.1 & 65.5/65.2 & 60.8/-& 70.0/69.0 &69.6/69.3 & \textbf{71.5}/66.9& \underline{70.5}/\textbf{76.9}\\
		\cline{2-12}
		&BraTS & 66.3/94.8 & 60.6/94.5 & 71.2/95.9 & 60.0/87.3& 67.3/93.2 & 76.9/- & 84.6/96.1  &\underline{86.8}/\underline{97.3}& 84.9/97.1& \textbf{90.5}/\textbf{97.9} \\
		\hline
		\multirow{6}{*}{8}& VisA & 73.7/94.2 & 77.9/95.0 & 85.4/93.9 & 80.0/95.3 & 85.4/95.4 & 88.7/-& 92.3/\underline{97.8}&\textbf{94.0}/97.5&\underline{93.4}/97.7   & 93.0/\textbf{98.6} \\
		& MVTecAD & 78.9/95.7 & 85.0/95.6 & 90.2/94.8 & 88.2/95.9 & 94.8/94.5 & 95.3/-&  97.7/96.7&97.1/95.0 & \textbf{98.1}/\underline{97.1} & \underline{97.8}/\textbf{97.6} \\
		& BTAD  & 84.2/96.3 & 93.2/\underline{97.2} & 91.4/96.0  & 91.6/\textbf{97.3} & 90.2/96.0 & 89.0/- & 91.6/96.8 &93.5/96.9& \textbf{95.3}/96.4 & \underline{94.6}/97.0\\
		& MVTec3D & 63.8/95.6 & 63.5/96.3 & 68.3/94.2 & 67.4/96.9 & 75.8/97.1 & 71.4/- & 83.0/97.9  &\underline{86.0}/98.1& \underline{86.0}/\textbf{98.9} & \textbf{89.0}/\underline{98.6}\\
		
		&MVTecLOCO & 64.3/67.3 & 57.7/69.1 & 65.7/\underline{69.7} & 62.2/69.6 & 68.0/65.5 & 62.9/- & 69.5/69.3  &\underline{71.2}/\underline{69.7}& 69.7/68.1  & \textbf{71.7}/\textbf{76.7}\\
		
		\cline{2-12}
		&BraTS & 72.6/95.4 & 71.2/96.0 & 76.4/96.4 &66.6/86.1
		& 68.9/93.8 & 79.3/- & 85.9/96.3 &\underline{88.7}/97.6& 88.5/\underline{98.0}& \textbf{90.1}/\textbf{98.1}\\
		\hline
	\end{tabular}
}
	\vspace{+0.1 cm}
\end{table}


\begin{table}
	\centering
	\vspace{-0.1 cm}
	\begin{minipage}[t]{0.50\textwidth}
		\centering
		\vspace{-0.2 cm}
		\vtop{
			\vspace{0pt}
			\centering
			\includegraphics[width=0.999\columnwidth]{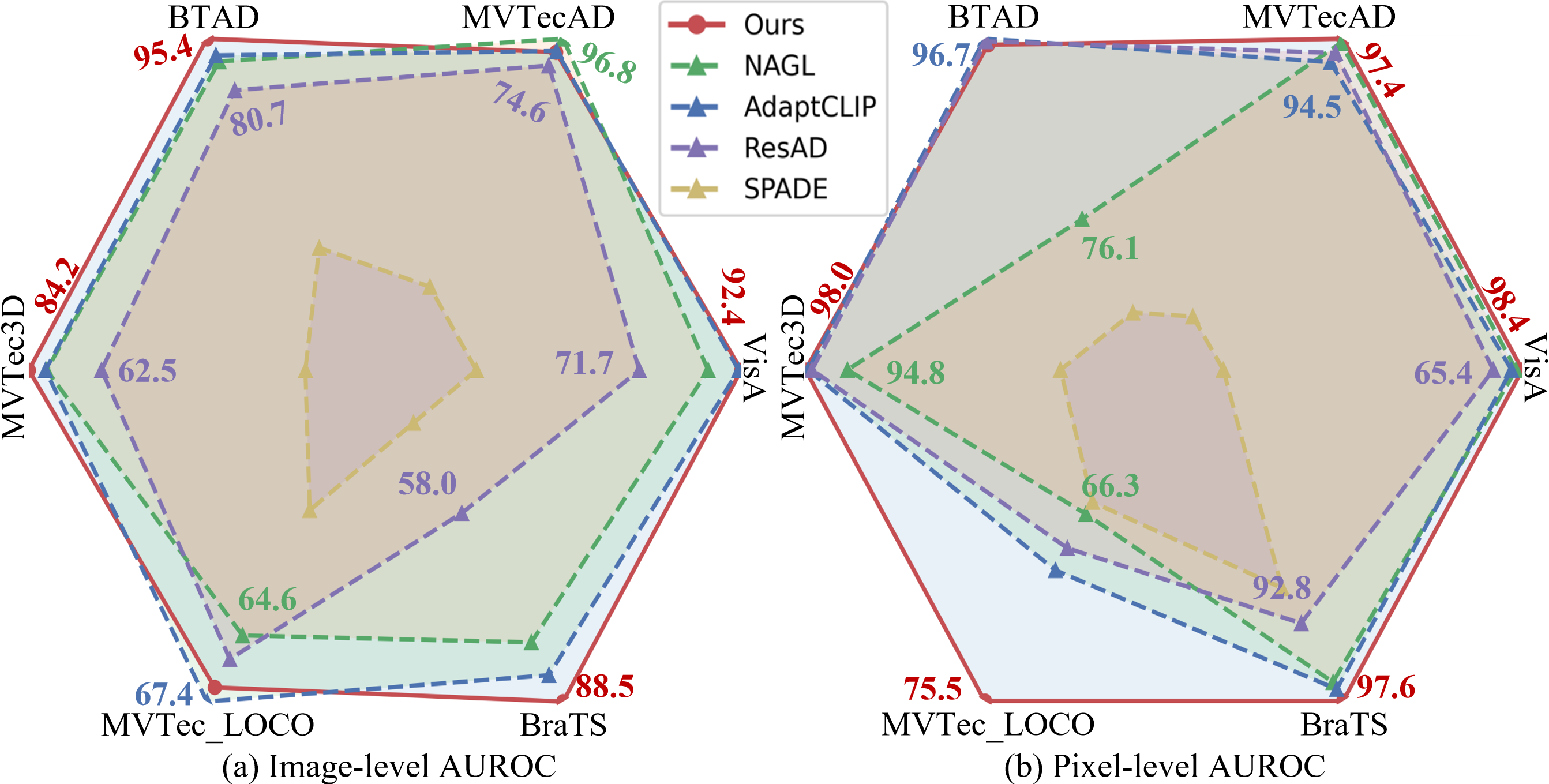}
		}
		\vspace{+0.1cm}
		\captionof{figure}{2-shot performance comparison across different datasets. Results are normalized per dataset for better visualization.}\label{fig:radar}
		\vspace{-0.1cm}
		\label{fig:tsne}
	\end{minipage}
	\begin{minipage}[t]{0.48\textwidth}
		\centering
		\vspace{-0.5 cm}
		\caption{
			VOD with more complex backbones.
		}\label{tab:backbone}
		\vspace{-0.03cm}
		\renewcommand{\arraystretch}{0.95}
		\resizebox{1\hsize}{!}{
			\setlength{\tabcolsep}{6pt}
			\begin{tabular}{c|ccc}
				\toprule
				Dataset & DINOv2-S & DINOv2-B & DINOv2-L \\
				\midrule
				VISA & \textbf{92.4}/\textbf{98.4}& \textbf{92.4}/97.7 & 92.2/97.9 \\
				MVTecAD & 95.7/\textbf{97.4}& \textbf{96.0}/97.1 & 95.8/96.3 \\
				BTAD & 95.4/96.0 & 95.1/97.0 & \textbf{95.9}/\textbf{97.3} \\
				MVTecLOCO & 66.7/\textbf{75.5}& 68.3/75.0 & \textbf{71.8}/74.5 \\
				BraTS & 88.5/97.6 & 89.2/98.2 & \textbf{90.7}/\textbf{98.8} \\
				\bottomrule
			\end{tabular}
		}
		\vspace{-0.2cm}
		\centering
		\caption{
			VOD with synthetic/real $\Ac$.
			\label{tab:real_a_refer}
		}
		\vspace{-0.03cm}
		\renewcommand{\arraystretch}{0.95}
		\resizebox{1\hsize}{!}{
			\setlength{\tabcolsep}{3pt}
			\begin{tabular}{cc|cccccc}
				\toprule
				\multicolumn{2}{c|}{VOD}& VisA & {\makecell{MVTec\\[-3pt] AD}} & BTAD & \makecell{MVTec\\[-3pt] 3D} & \makecell{MVTec\\[-3pt] LOCO} &BraTS\\
				\midrule
				\multirow{2}{*}{Ima.}&real&\textbf{93.3}&96.5&\textbf{95.0}&\textbf{85.3}&\textbf{70.6}&90.4 \\
				&syn.&\textbf{93.3}&\textbf{96.8}&\textbf{95.0}&85.0&70.5&\textbf{90.5} \\
				\hline
				\multirow{2}{*}{Pix.}&real&\textbf{98.5}&\textbf{97.5}&96.7&98.3&\textbf{77.0}&\textbf{98.0} \\
				&syn.&\textbf{98.5}&\textbf{97.5}&\textbf{96.8}&\textbf{98.7}&76.9&97.9 \\
				\bottomrule
			\end{tabular}
		}
	\end{minipage}
	\vspace{-0.5cm}
\end{table}

To demonstrate the strong domain generalization ability under limited source data (\ie few-shot normal references), we follow the one-to-many setting \cite{wang2025normalabnormal}, training on MVTecAD and evaluating on other datasets, and vice versa for MVTecAD using VisA as the training set.

\textbf{Quantitative comparison.}
The quantitative results of our VOD across different datasets under different few-shot settings are summarized in Table \ref{tab:one2many}.
Although trained using synthetic defects, our method generalizes effectively to real defect detection and consistently outperforms methods that either do not use reference images or rely solely on normal references across all shot settings (except for a few cases where it is surpassed by ResAD, RegAD or AdaptCLIP). Notably, our method is even comparable to, and often surpasses, NAGL, which relies on real abnormal references.
When extended to the medical dataset BraTS, our method outperforms other methods by a large margin, further verifying its effectiveness in mitigating domain gaps.
Figure \ref{fig:radar} presents a radar plot comparing selected competing methods. In terms of image-level AUROC, our method performs comparably to AdaptCLIP and outperforms other methods by a large margin on most datasets. In terms of pixel-level AUROC, our method consistently achieves the best performance across most datasets, with particularly significant improvements on MVTecLOCO. Although NAGL demonstrates competitive performance in image-level AUROC, it yields the second-lowest pixel-level AUROC in most cases.

As different datasets involve diverse object categories and defect types, the superior performance across multiple industrial datasets also demonstrates that VOD effectively mitigates the gaps across both object categories and defect types.

\begin{figure}[t]
	\vspace{-0.28 cm}
	\setlength{\abovecaptionskip}{2.0pt}
	\centering
	\includegraphics[width=0.99\columnwidth]{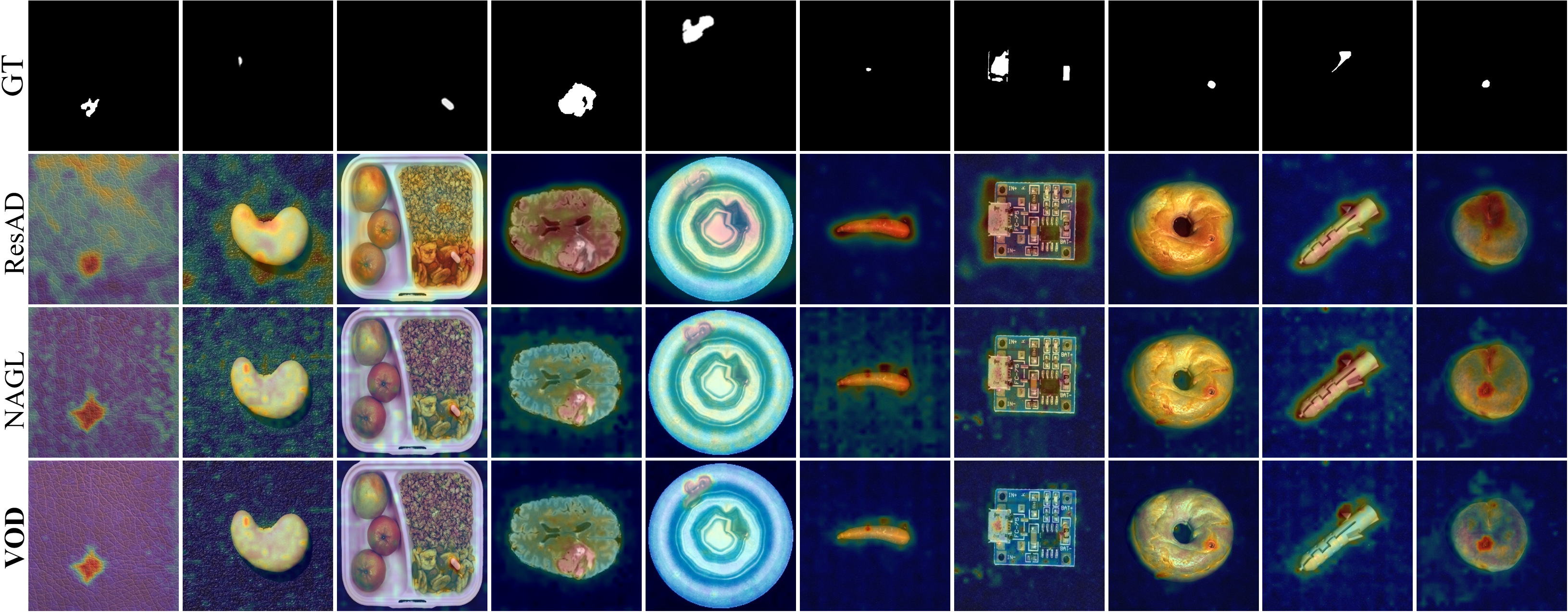}
	\vspace{-0.03 cm}
	\caption{Qualitative Comparison of anomaly detection results across different datasets.
	}\label{fig:maps_mask}
	\vspace{-0.5 cm}
\end{figure}

\textbf{Qualitative comparison.}
Several straightforward visualization results from industrial and medical datasets under the 4-shot setting are shown in Figure \ref{fig:maps_mask}.
As can be observed, ResAD exhibits false alarms in anomaly localization when generalizing. NAGL produces noisy predictions in the background, \eg assigning relatively high responses to background regions in the 1st, 2nd, and last cases.
In comparison, our method achieves more accurate anomaly localization and is less affected by background noise.

\subsection{Effect of backbone complexity and real abnormal references}

We further analyze performance from two perspectives: ($i$) increasing backbone capacity under the 2-shot setting (\ie using DINOv2-Base or DINOv2-Large instead of DINOv2-Small), and ($ii$) replacing synthetic abnormal references with real ones under the 4-shot setting. Results are reported in Tables \ref{tab:backbone} and \ref{tab:real_a_refer}.
We observe that: ($i$) as the backbone complexity increases, the performance improves steadily in most cases, with DINOv2-Large achieving the best results at the cost of additional computational overhead; 
($ii$) incorporating real abnormal references does not lead to noticeable performance gains across all test datasets, which further supports the effectiveness of our proposed value space in enabling synthetic abnormalities to serve as reliable proxies for real ones.

\subsection{Ablation Study}

\begin{table*}[t]
	\vspace{-0.8 cm}
	\centering
	\caption{
	Ablation studies. \ding{55} indicates that the corresponding component is removed.
		\label{tab:Ablation}
	}
	\vspace{-0.26 cm}
	\renewcommand{\arraystretch}{1.0}
	\resizebox{0.99\hsize}{!}{
		\setlength{\tabcolsep}{3pt}
		\begin{tabular}{ccccc|cccccc}
			\toprule
			\multirow{2}{*}{}
			\multirow{2}{*}{} & \multirow{2}{*}{} & \multirow{2}{*}{} & \multirow{2}{*}{} & \multirow{2}{*}{}
			& \multicolumn{6}{c}{\textbf{Datasets}} \\
			& Adaptation & VOD & $\Smat^{\tn}$ & $\hat{\Smat^{\ta}}$ & VisA & MVTecAD & BTAD & MVTec3D & MVTecLOCO &BraTS\\
			\midrule
			\romannumeral 1 &\ding{55}&&&&87.5/96.7&95.2/96.6&89.5/96.2&80.7/97.3&68.6/76.4&\textbf{97.2}/\textbf{97.9} \\
			\romannumeral 2 &&\ding{55}&&& 90.1/97.2&94.7/96.8&93.4/93.4&81.2/97.6 &69.4/\textbf{77.1}& 81.9/96.2 \\
			\romannumeral 3 &&&\ding{55} && \textbf{93.4}/\textbf{98.5}&96.6/\textbf{97.6}&91.5/93.4&\underline{84.7}/92.2&\textbf{70.8}/\underline{76.9}&89.9/\underline{97.8}  \\
			\romannumeral 4 &&&&\ding{55} &90.5/\underline{97.6} &\underline{96.7}/97.4&\underline{94.6}/\underline{96.6} & 83.5/\underline{97.8}&69.6/76.8&85.8/97.7 \\
			\romannumeral 5 &&&&& \underline{93.3}/\textbf{98.5}&\textbf{97.8}/\underline{97.5}&\textbf{95.1}/\textbf{96.8}&\textbf{85.0}/\textbf{98.7}&\underline{70.5}/\underline{76.9}&\underline{90.5}/\textbf{97.9} \\
			
			\bottomrule
		\end{tabular}
	}
	\vspace{-0.1 cm}
\end{table*}

To evaluate the individual contribution of each proposed component to the overall performance, we remove each component in turn and report the results in Table \ref{tab:Ablation} (see Appendix \ref{app:sn_sa} for more results).
($i$) Without adaptation (row i), both feature representation and separability are entirely determined by the pre-trained model, leading to the lowest performance among all variants. An exception is observed on BraTS, where the model without adaptation unexpectedly outperforms its adapted counterparts. We attribute this to the domain bias introduced by adapting on MVTecAD, which shifts the model toward the industrial domain and consequently degrades its generalization to the medical domain.
($ii$) When adaptation is enabled (row ii), VOD contributes more significantly to performance improvements than $\Smat^{\tn}$ and $\hat{\Smat^{\ta}}$ (as observed on VisA, MVTecAD, MVTec3D, and BraTS). This result highlights the importance of aligning both normal and abnormal distributions across the three generalization gaps via value-order decomposition.
($iii$) Comparing row iii and iv,
$\hat{\Smat^{\ta}}$ has a larger impact than $\Smat^{\tn}$ on the final performance (see ima. AUROC on VisA, MVTec3D, MVTecLOCO, and BraTS), demonstrating the importance of incorporating synthetic abnormal samples. 


\section{Conclusions}
\label{sec:conclusions}


With the goal of GAD, we propose VOD, a simple yet effective method trained on a single source dataset and generalizes to unseen target domains. We identify three generalization gaps across object categories, defect types, and data domains, and disentangle representations into two components: value and order. The value component serves as a gap-invariant indicator of abnormality, particularly robust across real and synthetic anomalies, while the order component captures gap-specific variations.
This disentanglement may provide insights for more robust and generalizable anomaly detection.
Based on the value component, we build a VOD-based anomaly detection framework using only a few-shot set of normal references. Experimental results show consistent improvements over existing methods in cross-domain generalization.
Despite its effectiveness, VOD relies on the diversity and quantity of normal references, which is inherent to reference-based methods. Extending it to logical anomaly detection requires modeling complex global structures, which we leave for future work.
This work improves the reliability of industrial systems through early anomaly detection, but incorrect predictions may still introduce operational risks, highlighting the need for careful validation and human oversight in practical deployment.

%
%

{
	\small
	\bibliographystyle{unsrt}
	\bibliography{ReferencesCong}
}



\newpage

\appendix{}

Appendix for ``Value-order Decomposition for Generalist Anomaly Detection''

\section{Datasets}

The details of the datasets used in this work are summerized in Table \ref{tab:data_info}

\begin{table}[h]
	\centering
	\caption{
			Statistics of the datasets used in this work.
		}
	\label{tab:data_info}
	\renewcommand{\arraystretch}{1.5}
	\resizebox{0.99\hsize}{!}{
		\setlength{\tabcolsep}{6pt}
		\begin{tabular}{c|cccc}
			\toprule
			Data name & \# Object categories & \# Defect types & \# Normal test samples & \# Abnormal test samples \\
			\midrule
			MVTecAD & 15 &  73 & 467 & 1258\\
			VisA & 12 & 1 & 962 & 1200  \\
			BTAD & 3  & 1 & 451 & 290 \\
			MVTec3D &  10 & 40 & 249 & 948 \\
			MVTecLOCO & 5  & 10 & 575 & 993 \\
			BraTS & 1 & 1 & 154 & 1097 \\
			\bottomrule
		\end{tabular}
	}
\end{table}

\textbf{MVTecAD \cite{bergmann2019mvtec}}.
MVTecAD is a widely used benchmark for unsupervised anomaly detection. It contains 5,354 high-resolution RGB images from 15 categories, including both textures (5 categories) and objects (10 categories). The training set consists of 3,629 normal images, while the test set includes 1,725 samples with both normal and anomalous instances. The anomalies span more than 70 defect types, such as scratches, crack, contamination, and structural irregularities.

\textbf{VisA \cite{zou2022spot}}.
VisA is another popular benchmark dataset for anomaly detection, comprising 10,821 high-resolution color images. It includes 9,621 normal samples and 1,200 anomalous samples across 12 object categories in three different domains. The dataset is particularly challenging due to its diverse and complex visual conditions, including industrial PCBs with intricate layouts, multiple objects with significant variations in pose and position, and loosely aligned instances with subtle appearance differences. The anomalies involve a wide range of defects, covering both surface-level and structural variations.

\textbf{BTAD \cite{9576231}}.
BTAD is designed for multiple vision tasks, including instance segmentation, semantic segmentation, and object detection. It consists of real-world industrial images containing both surface and structural defects.

\textbf{MVTec3D \cite{bergmann2021mvtec}}.
MVTec3D is a dataset for unsupervised anomaly detection and localization in 3D industrial scenarios, where certain defects manifest as geometric inconsistencies. It provides 4,147 point cloud scans along with corresponding 2D RGB images across 10 categories. As our method operates on 2D inputs, we only utilize the RGB images.

\textbf{MVTecLOCO \cite{bergmann2022beyond}}.
MVTecLOCO focuses on unsupervised anomaly detection involving both structural and logical anomalies in industrial inspection tasks. The dataset includes 1,772 training images, 304 validation images, and 1,568 test images across five categories, each defined by specific logical rules. For example, a breakfast box is required to contain exactly two tangerines and one nectarine arranged on the left side, while a screw bag must include two washers, two nuts, one long screw, and one short screw. Any deviation from these rules is considered a logical anomaly.

\textbf{BraTS \cite{6975210}}.
BraTS is a well-established multimodal MRI dataset for brain tumor segmentation, released annually as part of the MICCAI challenges since 2012. In this work, we adopt the 2021 BraTS dataset, which includes 1,251 labeled cases, comprising 1,097 tumor cases and 154 normal cases without tumors.

\section{Experimental setup}
\label{app:setup}

Given the few-shot normal references, we adopt the synthetic anomaly generation strategy from DRAEM \cite{zavrtanik2021draem}. Specifically, for each normal reference, we first generate a noise image using a Perlin noise generator to model diverse anomaly shapes, and then binarize it with a threshold such that only 20\% of the values remain non-zero, forming a mask $M$. We randomly sample a texture image from DTD \cite{cimpoi2014describing} as the anomaly source. To further enhance the distinctiveness of the anomalous regions, we apply random augmentation by sampling three operations from the set $\{\text{posterize, sharpness, solarize, equalize, brightness change, color change, auto\text{-}contrast}\}$.

\textbf{Implementation Details.} We adopt a DINOv2 \cite{oquab2024dinov} pre-trained Vision Transformer (ViT) as the feature extractor and freeze all its parameters during training. 
For the adaptation block, we adopt a Transformer encoder layer with 4 attention heads to improve the discriminability between normal and abnormal features. The block consists of a feedforward network with an expansion dimension of 1536, uses GELU activation to enhance representational capacity, and applies a dropout rate of 0.1 for regularization.
The backbones used by the compared methods are summarized in Table \ref{tab:backbone_c}. 

Our method is trained for 20 epochs, with each epoch consisting of 500 sampled episodes and a batch size of 8. We use the AdamW optimizer with a weight decay of $1\times 10^{-5}$. The initial learning rate is set to $1\times 10^{-5}$ and decayed by a factor of 0.1 at epochs 10 and 15. All training and testing images are resized to $448 \times 448$, and all experiments are conducted on a single NVIDIA A40 GPU.
Following prior work \cite{yao2024resad,yao2025resadpp}, we train the model on one dataset and evaluate it on the others. The number of few-shot normal reference samples is set to $K\in\{2,4,8\}$. Our implementation is based on the publicly released codebase of NAGL \cite{wang2025normalabnormal}, and we follow the same model selection protocol, where the validation set is constructed from VisA. We thank the authors for their contribution to the community.

%

\begin{table}[h]
	\centering
	\caption{
		Backbone and parameter size of each compared method.
		\label{tab:backbone_c}
	}
	\renewcommand{\arraystretch}{1.5}
	\resizebox{1\hsize}{!}{
		\setlength{\tabcolsep}{3pt}
		
		\begin{tabular}{c|cccccccccc}
			\toprule
			Methods & SPADE & PaDiM & Patchcore & RegAD & WinCLIP&InCTRL & ResAD & AdaptCLIP & NAGL & \textbf{VOD}  \\
			\midrule
			Backbones&  \multicolumn{3}{c}{WideResNet50-2} & ResNet-18&  \multicolumn{2}{c}{ViT-B/16+} & ImageBind-Huge &  ViT-L/14@336px & \multicolumn{2}{c}{DINOv2-Small} \\
			Paramters&  \multicolumn{3}{c}{68.9 M} & 11.7 M &\multicolumn{2}{c}{86.6 M} &2.3 B&303.9 M&\multicolumn{2}{c}{21.7 M} \\
			\bottomrule
		\end{tabular}
	}
\end{table}

\section{Additional experimental results}
\label{sec:app:moreE}

\subsection{Contributions of the similarity scores $\Smat^{\tn}$ and $\hat{\Smat^{\ta}}$}
\label{app:sn_sa}

\begin{figure}[h]
	\vspace{-0.2 cm}
	\setlength{\abovecaptionskip}{2.0pt}
	\centering
	\includegraphics[width=0.99\columnwidth]{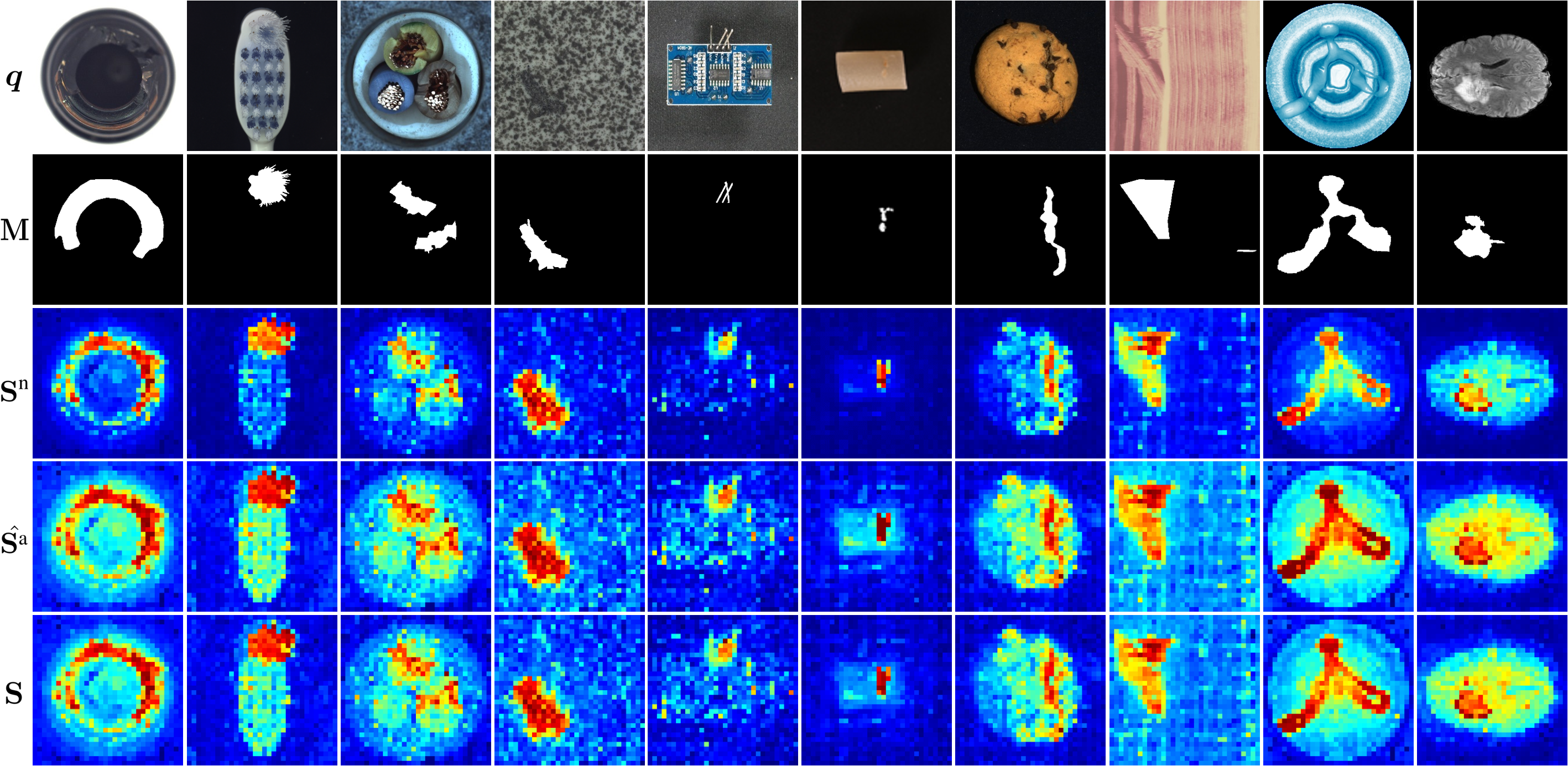}
	\vspace{-0.0 cm}
	\caption{Visualize the contribution of the similarity scores $\Smat^{\tn}$ and $\hat{\Smat^{\ta}}$.
	}\label{fig:sn_sa}
\end{figure}

To better understand the contributions of the similarity scores $\Smat^{\tn}$ and $\hat{\Smat^{\ta}}$, we further visualize them, along with their combined anomaly score $\Smat$, in Figure \ref{fig:sn_sa}.
In most cases, $\Smat^{\tn}$ is able to localize the most salient abnormal regions but tends to miss less prominent ones, whereas $\hat{\Smat^{\ta}}$ provides more precise localization of fine-grained abnormal details. By combining the two, the final anomaly segmentation results are more accurately aligned with the ground truth.

\subsection{Efficiency analysis}
\label{sec:app:efficiency}

All experiments are conducted on a single NVIDIA A40 GPU. We compare the efficiency of our method with NAGL, AdaptCLIP, and ResAD in terms of the total number of parameters, training time and inference speed.

As shown in Table \ref{tab:efficiency}, the parameter size has a significant impact on training time and inference speed. For example, ResAD and AdaptCLIP have substantially larger parameter sizes than NAGL and VOD, and thus exhibit lower efficiency.
Our method introduces slightly fewer parameters than NAGL, while achieving a slightly lower inference speed. This is due to the additional process of generating synthetic abnormal references.
Overall, VOD achieves a favorable balance between efficiency and performance.

\begin{table*}[h]
	\centering
	\caption{
			The efficiency comparison under the 4-shots setting.
		}
	\label{tab:efficiency}
	\renewcommand{\arraystretch}{1.1}
	\resizebox{0.9\hsize}{!}{
			\setlength{\tabcolsep}{7pt}
			\begin{tabular}{c|rrr}
					\toprule
					Method & Total Parameters (M) & Training Time (H) & Inference Speed (FPS) \\
					\midrule
					ResAD & 59.2 & 20.6 & 7.8 \\
					AdaptCLIP & 429.7 & 1.5 & 5.9  \\ 
					NAGL & 24.4 & 0.3 & 17.1  \\
					\textbf{VOD} & 23.8 & 0.6 &  14.3  \\
					\bottomrule
				\end{tabular}
		}
\end{table*}


\subsection{The effect of the hyperparameter $\lambda$}
\label{sec:app:hyperpara}

\begin{figure}[h]
	\setlength{\abovecaptionskip}{2.0pt}
	\centering
	\includegraphics[width=0.9\columnwidth]{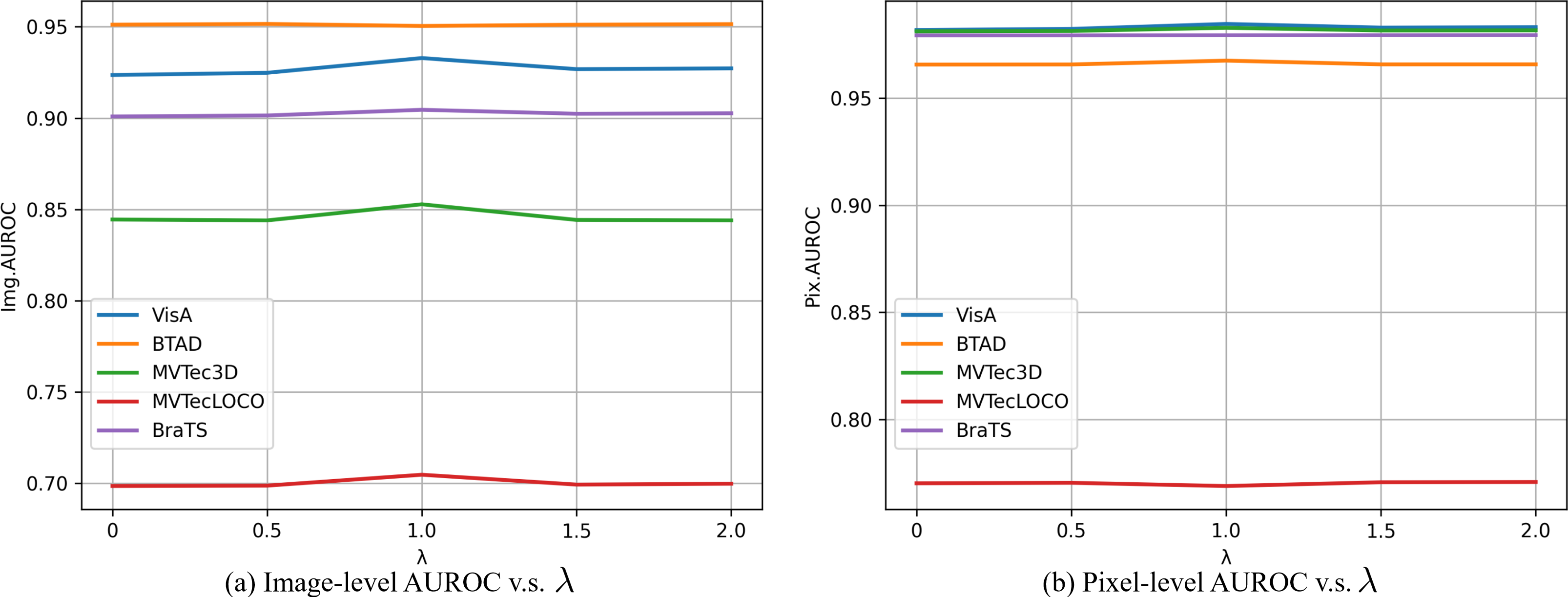}
	\caption{The effect of $\lambda$ on our VOD.
	}\label{fig:lambda}
\end{figure}

The hyperparameter $\lambda$ is used to balance the image-level and pixel-level prediction losses. We analyze its effect on the final performance by varying $\lambda$ from 0 to 2 with a step size of 0.5, where $\lambda = 0$ corresponds to using only the image-level prediction loss. The results, in terms of image-level and pixel-level AUROC across five test datasets, are shown in Figure \ref{fig:lambda}. We observe that our chosen setting $\lambda = 1$ consistently achieves the best performance on both metrics across all datasets, highlighting the importance of jointly considering image-level and pixel-level supervision.

\subsection{Results with more metrics}

To provide a more comprehensive evaluation of VOD, we further adopt additional metrics following \cite{wang2025normalabnormal, gao2026adaptclip}. For image-level anomaly detection, we report the Area Under the Receiver Operating Characteristic Curve (AUROC), the maximum F1-score at the optimal threshold (F1-max), and Average Precision (AP). For pixel-level anomaly localization, we additionally employ the per-region overlap (PRO) metric together with AUROC and F1-max.
Each metric is explained as follows.

\begin{itemize}
	\item \textbf{AUROC.} AUROC measures the area under the Receiver Operating Characteristic (ROC) curve and evaluates the overall ranking ability of anomaly scores across all thresholds. A higher AUROC indicates better separability between normal and anomalous samples. For pixel-level evaluation, it may be biased toward large anomalous regions since larger defects contribute more pixels.
	
	\item \textbf{F1-max.} F1-max is the maximum F1-score achieved over all possible decision thresholds, reflecting the best achievable detection performance under optimal threshold selection.
	
	\item \textbf{AP.} Average Precision (AP) is the area under the Precision\text{-}Recall (PR) curve, which evaluates the overall ability of the model to detect anomalous samples by balancing precision and recall across all thresholds. It is particularly suitable for anomaly detection tasks with severe class imbalance.
	
	\item \textbf{PRO.} Per-Region Overlap (PRO) measures localization quality at the region level by computing the average correctly classified pixel ratio for each connected anomalous region under different false positive rates (typically within $[0, 0.3]$). The PRO score is defined as the normalized area under this curve. In this work, we report the PRO score with false positive rates as 0.3. A higher PRO indicates better localization performance for both large and small anomalies.
	
\end{itemize}


We compare VOD with existing methods on additional image-level (AUROC, AP, and F1-max) and pixel-level (AUROC, PRO, and F1-max) metrics. 
The results on MVTec and VisA under 2-shot and 4-shot settings are summarized in Table \ref{tab:app:other_metric}.
Overall, VOD achieves the best performance on the majority of metrics across both datasets and shot settings. 
Notably, even compared with NAGL, which leverages real anomaly samples as references, VOD outperforms it in most cases.
Additionally, as the number of reference shots increases, VOD consistently achieves improved performance across all metrics, highlighting its strong scalability.

\begin{table}[h]
	\centering
	\caption{
		Comparison with existing methods on MVTec and VisA under 2-shot and 4-shot few shot settings.
		\label{tab:app:other_metric}
	}
	\renewcommand{\arraystretch}{1.0}
	\resizebox{1\hsize}{!}{
		\setlength{\tabcolsep}{3pt}
		
		\begin{tabular}{c|c|ccc|ccc|ccc|ccc}
			\toprule
			&&\multicolumn{6}{c|}{MVTecAD}&\multicolumn{6}{c}{VisA}\\
			\hline
			&&\multicolumn{3}{c|}{Ima.}&\multicolumn{3}{c|}{Pix.}&\multicolumn{3}{c|}{Ima.}&\multicolumn{3}{c}{Pix.}\\
			Shots&Methods& AUROC&AP&F1-max&AUROC&PRO&F1-max& AUROC&AP&F1-max&AUROC&PRO&F1-max\\
			\midrule
			\multirow{7}{*}{2}
			&SPADE&82.9&91.7&91.1&92.0&85.7&44.5&80.7&82.3&81.7&96.2&85.7&40.5\\
			&PatchCore&86.3&93.8&92.0&93.3&82.3&53.0&81.6&84.8&82.5&96.1&82.6&41.0\\
			&WinCLIP&94.4&97.0&94.4&96.0&88.4&58.4&84.6&85.8&83.0&96.8&86.2&43.5\\
			&ResAD&87.2&93.9&92.2&94.8&85.5&50.2&86.6&88.3&84.1&96.5&82.3&40.0\\
			&AdaptCLIP&95.7&97.9&95.4&94.5&90.5&54.9&92.3&\textbf{94.4}&\textbf{88.8}&97.1&91.8&45.6\\
			&NAGL&\textbf{96.8}&97.9&96.3&96.8&93.2&\textbf{59.8}&89.8&90.6&86.8&97.6&91.4&43.3\\
			&\textbf{VOD}&95.7&\textbf{98.1}&\textbf{96.7}&\textbf{97.4}&\textbf{93.7}&59.5&\textbf{92.4}&92.5&88.4&\textbf{98.4}&\textbf{92.1}&\textbf{46.0}\\
			\midrule
			\multirow{7}{*}{4}
			&SPADE&84.8&92.5&91.5&92.7&87.0&46.2&81.7&83.4&82.1&96.6&87.3&43.6\\
			&PatchCore&88.8&94.5&92.6&94.3&84.3&55.0&85.3&87.5&84.3&96.8&84.9&43.9\\
			&WinCLIP&95.2&97.3&94.7&96.2&89.0&59.5&87.3&88.8&84.2&97.2&87.6&\textbf{47.0}\\
			&ResAD&90.7&95.7&93.9&95.8&88.7&53.0&89.3&90.7&86.5&96.8&84.1&41.6\\
			&AdaptCLIP&96.6&98.3&96.0&94.8&90.8&56.1&93.1&\textbf{94.5}&88.8&97.3&\textbf{92.2}&46.7\\
			&NAGL&\textbf{97.1}&98.0&96.4&97.0&93.5&60.1&91.2&91.7&87.8&97.8&91.5&44.0\\
			&\textbf{VOD}&96.8&\textbf{98.7}&\textbf{96.9}&\textbf{97.5}&\textbf{94.0}&\textbf{60.6}&\textbf{93.3}&93.5&\textbf{89.2}&\textbf{98.5}&\textbf{92.2}&46.1\\
			\bottomrule
			
		\end{tabular}
	}
\end{table}

We additionally report the results of VOD under 2/4/8-shot settings for all datasets in Table \ref{tab:app:2shot}, \ref{tab:app:4shot}, and \ref{tab:app:8shot} for completeness.

\begin{table}[h]
	\centering
	\caption{
		VOD results with additional metrics under the 2-shot setting.
		\label{tab:app:2shot}
	}
	\vspace{-0.03cm}
	\renewcommand{\arraystretch}{1.1}
	\resizebox{0.9\hsize}{!}{
		\setlength{\tabcolsep}{6pt}
		
		\begin{tabular}{cc|cccccc}
			\toprule
			& & \multicolumn{6}{c}{Datasets} \\
			\multicolumn{2}{c|}{Metrics} & VisA & MVTecAD & BTAD & MVTec3D & MVTecLOCO & BraTS \\
			\midrule
			\multirow{3}{*}{Ima.}&AUROC&92.4&95.7& 95.4& 82.4 & 66.7 & 88.5 \\
			&AP&92.5& 98.1& 97.5& 94.8 & 77.6 & 98.0 \\
			&F1&88.4& 96.7& 93.6& 91.4 & 78.6 & 94.5 \\
			\midrule
			\multirow{3}{*}{Pix.}&PRO&92.1& 93.7 & 76.6& 93.7 & 66.1 & 81.8 \\
			&AUROC&98.4&97.4& 96.0& 98.0 & 75.5 & 97.6 \\
			&F1&46.0& 59.5 & 56.4& 50.1 & 25.2 & 57.5 \\
			\bottomrule
		\end{tabular}
	}
\end{table}

\begin{table}[h]
	\vspace{-0.5cm}
	\centering
	\caption{
		VOD results with additional metrics under the 4-shot setting.
		\label{tab:app:4shot}
	}
	\vspace{-0.03cm}
	\renewcommand{\arraystretch}{1.1}
	\resizebox{0.9\hsize}{!}{
		\setlength{\tabcolsep}{6pt}
		
		\begin{tabular}{cc|cccccc}
			\toprule
			& & \multicolumn{6}{c}{Datasets} \\
			\multicolumn{2}{c|}{Metrics} & VisA & MVTecAD & BTAD & MVTec3D & MVTecLOCO & BraTS \\
			\midrule
			\multirow{3}{*}{Ima.}&AUROC& 93.3 & 96.8 & 95.0 & 85.0 & 70.5 & 90.5 \\
			&AP& 93.5 & 98.7 & 97.5 & 95.5 & 80.1 & 98.3 \\
			&F1& 89.2 & 96.9 & 93.6 & 92.6 & 79.1 & 94.8 \\
			\midrule
			\multirow{3}{*}{Pix.}&PRO& 92.2 & 94.0 & 74.5 & 95.9 & 66.0 & 82.9 \\
			&AUROC& 98.5 & 97.5 & 96.8 & 98.7 & 76.9 & 97.9 \\
			&F1& 46.1 & 60.6 & 58.4 & 52.2 & 27.0 & 60.8 \\
			\bottomrule
		\end{tabular}
	}
\end{table}

\begin{table}[h]
	\centering
	\caption{
		VOD results with additional metrics under the 8-shot setting.
		\label{tab:app:8shot}
	}
	\vspace{-0.03cm}
	\renewcommand{\arraystretch}{1.1}
	\resizebox{0.9\hsize}{!}{
		\setlength{\tabcolsep}{6pt}
		
		\begin{tabular}{cc|cccccc}
			\toprule
			& & \multicolumn{6}{c}{Datasets} \\
			\multicolumn{2}{c|}{Metrics} & VisA & MVTecAD & BTAD & MVTec3D & MVTecLOCO & BraTS \\
			\midrule
			\multirow{3}{*}{Ima.}&AUROC& 93.0 & 97.8 & 94.6 & 89.0 & 71.1 & 90.1 \\
			&AP& 93.0 & 98.9 & 97.3 & 96.9 & 81.3 & 98.1 \\
			&F1& 89.0 & 97.3 & 92.7 & 93.0 & 79.4 & 95.0 \\
			\midrule
			\multirow{3}{*}{Pix.}&PRO& 92.7 & 94.2 & 75.9 & 95.2 & 66.2 & 83.5 \\
			&AUROC& 98.6 & 97.6 & 97.0 & 98.6 & 76.7 & 98.1 \\
			&F1& 45.7 & 97.8 & 60.4 & 54.1 & 26.7 & 62.9 \\
			\bottomrule
		\end{tabular}
	}
\end{table}

\subsection{Full visualization of different representations across the three generalization gaps}
\label{sec:app:full_tsne}

\begin{figure}[h]
	\setlength{\abovecaptionskip}{2.0pt}
	\centering
	\includegraphics[width=0.999\columnwidth]{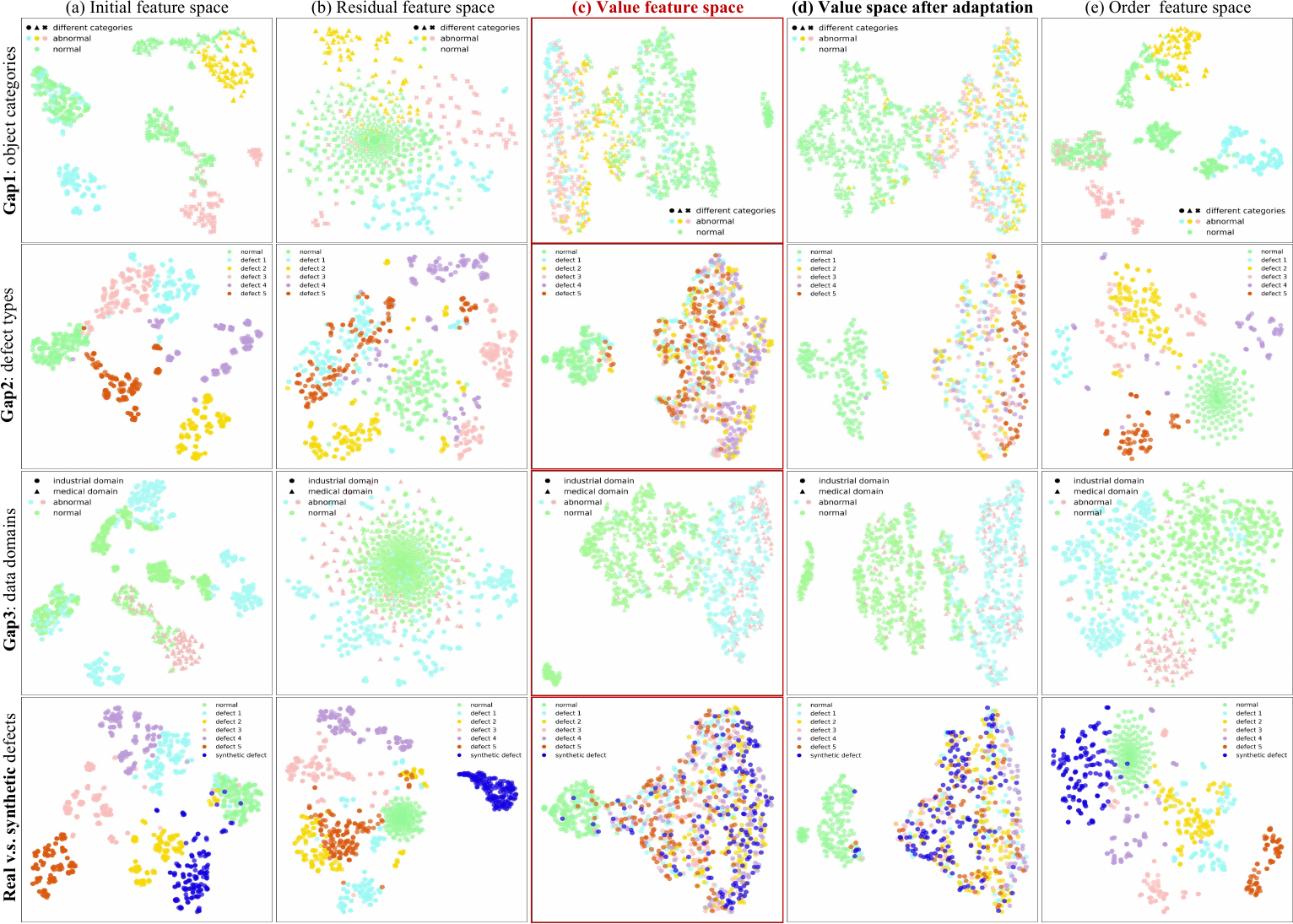}
	\vspace{-0.0 cm}
	\caption{t-SNE visualization of normal and anomalous patches under different representations.
		(a) The initial feature space exhibits three distinct gaps, which hinder generalization and negatively affect anomaly detection performance.
		(b) Residual-space methods mainly improve the alignment among normal samples.
		(c) In contrast, the value component of VOD enforces intra-class alignment for both normal and anomalous samples.
		(d) After adaptation, the value component further enhances the separability between normal and anomalous samples. 
		(e) The order component of VOD shows a similar variation across the three generalization gaps.
	}\label{fig:all_tsne}
	\vspace{+6.0 cm}
\end{figure}

A comprehensive visualization of normal and abnormal patch distributions under different representation spaces across the three generalization gaps is shown in Figure \ref{fig:all_tsne}. Each row is constructed as follows:

\begin{itemize}
	\item For \textbf{Gap1}, we select three objects (\ie bottle, carpet, and hazelnut) from the MVTecAD dataset. For each object, 150 image patches are randomly sampled from normal images, and another 150 patches are sampled from anomalous regions in abnormal images for visualization.
	\item For \textbf{Gap2}, we select one object (\ie carpet) with five defect types from the MVTecAD dataset. Specifically, 150 patches are randomly sampled from normal images, and for each defect type, 150 patches are randomly sampled from the anomaly regions based on the ground-truth masks.
	\item For \textbf{Gap3}, we select an industrial dataset (MVTecAD) and a medical dataset (BraTS) for demonstration. From MVTecAD, we choose three object categories (\ie bottle, carpet, and hazelnut), and for each category, 150 patches are randomly sampled from normal images and 150 from anomaly regions. For BraTS, which contains a single category (\ie brain), 150 patches are randomly sampled from normal images and 150 from lesion regions based on the provided masks.
	\item For \textbf{real vs. synthetic} defects, we generate synthetic anomalies on carpet images from the MVTecAD dataset. For each real defect type and synthetic anomaly type, 150 patches are randomly sampled from the corresponding anomaly regions. In addition, 150 patches are sampled from normal images.
\end{itemize}

\subsection{Discussion on the stability of the VOD performance}
\label{sec:app:random_seed}

We report AUROC comparison results with standard deviations over three random seeds across six test datasets for ResAD, NAGL, and VOD in Table \ref{tab:p_std}.
($i$) For image-level AUROC, NAGL exhibits relatively large variability, while both ResAD and our VOD show significantly lower deviations, with ResAD achieving the smallest variance.
($ii$) For pixel-level AUROC, both ResAD and NAGL demonstrate large deviations on certain datasets, whereas our VOD maintains consistently low variability across all datasets.
These results demonstrate the superior stability and robustness of our method with respect to different random seeds.

\begin{table}[h]
	\centering
	\caption{
		AUROC (\%) with mean and standard deviation over 3 random seeds.
		\label{tab:p_std}
	}
	\renewcommand{\arraystretch}{1.2}
	\resizebox{0.99\hsize}{!}{
		\setlength{\tabcolsep}{3pt}
		
		\begin{tabular}{cc|cccccc}
			\toprule
			& & \multicolumn{6}{c}{Datasets} \\
			\multicolumn{2}{c|}{ } & VisA & MVTecAD & BTAD & MVTec3D & MVTecLOCO & BraTS \\
			\midrule
			\multirow{3}{*}{Ima.}&ResAD& 86.5$\pm$\textbf{0.31} & 90.7$\pm$\textbf{0.21} & 95.4$\pm$\textbf{0.12} & 70.5$\pm$\textbf{0.29} & - & - \\
			&NAGL& 91.2$\pm$1.1 & 97.1$\pm$0.6 & 94.4$\pm$1.2 & 86.9$\pm$1.8 & 71.5$\pm$3.1 & 84.9$\pm$1.7 \\
			&VOD& 93.3$\pm$0.53 & 96.8$\pm$0.43 & 95.0$\pm$0.21 & 85.0$\pm$0.97 & 70.5$\pm$\textbf{0.70} & 90.5$\pm$\textbf{0.13} \\
			\midrule
			\multirow{3}{*}{Pix.}&ResAD& 97.5$\pm$0.08 & 95.9$\pm$0.12 & 97.6$\pm$0.05 & 97.2$\pm$0.09 & - & - \\
			&NAGL& 97.8$\pm$0.0 & 97.0$\pm$\textbf{0.0} & 76.4$\pm$0.9 & 95.4$\pm$\textbf{0.0} & 66.9$\pm$0.6 & 97.1$\pm$0.4 \\
			&VOD& 98.5$\pm$\textbf{0.004} & 97.5$\pm$0.07 & 96.8$\pm$\textbf{0.06} & 98.7$\pm$0.03 & 76.9$\pm$\textbf{0.24} & 97.9$\pm$\textbf{0.03} \\
			\bottomrule
		\end{tabular}
	}
\end{table}


\end{document}